\documentclass[conference]{IEEEtran}
\IEEEoverridecommandlockouts
\usepackage{cite}
\usepackage{amsmath,amssymb,amsfonts}
\usepackage{algorithmic}
\usepackage{graphicx}
\usepackage{textcomp}
\usepackage{xcolor}
\usepackage{adjustbox}
\usepackage{enumerate}

\def\BibTeX{{\rm B\kern-.05em{\sc i\kern-.025em b}\kern-.08em
    T\kern-.1667em\lower.7ex\hbox{E}\kern-.125emX}}
\begin{document}

\title{A.I. and Data-Driven Mobility at \\
Volkswagen Financial Services AG\\
{\footnotesize }
\thanks{\IEEEauthorrefmark{1}Equal contribution}
}

\author{
\IEEEauthorblockN{Shayan Jawed\IEEEauthorrefmark{1}, Mofassir ul Islam Arif\IEEEauthorrefmark{1},
 Ahmed Rashed\IEEEauthorrefmark{1}, Kiran Madhusudhanan\IEEEauthorrefmark{1}, \\ Shereen Elsayed\IEEEauthorrefmark{1},
 Mohsan Jameel\IEEEauthorrefmark{1}
 and
Lars Schmidt-Thieme}
\IEEEauthorblockA{Information Systems and Machine Learning Lab,
University of Hildesheim, Germany\\
\{shayan, 
mofassir,
ahmedrashed,
kiranmadhusud,
elsayed,
mohsan.jameel,
schmidt-thieme\}@ismll.uni-hildeshiem.de}
\\
\IEEEauthorblockN{Alexei Volk, 
Andre Hintsches, 
Marlies Kornfeld and Katrin Lange
}
\IEEEauthorblockA{Data, Analytics \& A.I Unit, Volkswagen Financial Services AG, Germany \\
\{Alexei.Volk, 
Andre.Hintsches, 
Marlies.Kornfeld, 
Katrin.Lange\}@vwfs.io}
}

\maketitle

\begin{abstract}
Machine learning is being widely adapted in industrial applications owing to the capabilities of commercially available hardware and rapidly advancing research.
Volkswagen Financial Services (VWFS), as a market leader in vehicle leasing services, aims to leverage existing proprietary data and the latest research to enhance existing and derive new business processes. The collaboration between Information Systems and Machine Learning Lab (ISMLL) and VWFS serves to realize this goal. In this paper, we propose methods in the fields of recommender systems, object detection, and forecasting that enable data-driven decisions for the vehicle life-cycle at VWFS.
\end{abstract}


\section{Introduction}

Recent advances in Machine learning have enabled intelligent data-driven business decisions. 
VWFS generates data in several modalities, which include, image, and relational data, that enable the application of modern machine learning techniques to improve their B2B and B2C processes. 
These include (I) `Asset return forecast' that anticipates the return condition of a leased vehicle, (II) `Return assessment' that seeks to automate appraisals, (III) `Recommender system' for B2B used vehicle auctions, (IV) `Next Best Offer' that recommends next vehicles (B2C) to restart the entire pipeline.



\section{Asset return forecast}
An important decision at the end of the leasing period of a vehicle is to identify `the next best use' of the returned vehicle. Typically, this process begins after the return of the vehicle to the manufacturer. The ability to anticipate the return condition of these leased vehicles beforehand would reduce the delay in reusing the vehicle for the next use case. The return date, return mileage and return damage are key targets required by VWFS to determine the future use of a leased vehicle. Our work, predicts these key targets using machine learning algorithms (ML) from the leasing contract data. 

The leasing contract data consists of vehicle, customer, and dealer information for all the contracts from 2015-2020. After preprocessing, the dataset consists of more than 400,000 records with 82 predictors and 4 targets. For this use case, the machine learning models must be able to make predictions chronologically. Accordingly, the data was split into train and test such that given all the previous data, the model predicts the targets for the contracts expected to end within the next six months. Additionally, the task of predicting target values for each contract is redundant and therefore we simplify the task into a classification problem by segregating each target into multiple levels ranging from 0 to 5.

Recent literature demonstrates the superior performance of neural networks and boosted trees for residual value forecasting \cite{b3}. Following \cite{b3}, we propose two new models (a Catboost \cite{b4}, and a custom Neural Network \cite{b3}) to solve the task. We also compare their performance against a simple baseline (Majority class). In our experiments, for a test period from 07/2019-12/2019, the models (in the introduced order) were able to achieve an F1 score of 0.52, 0.50, 0.39 for the return prediction target, 0.44, 0.43, 0.28 for the mileage target and 0.41, 0.43, 0.34 for damage target showing significant improvement of ML methods over the naive baseline.


\section{Leasing Vehicle Return Assessment }
The final stage in the life-cycle of a vehicle leased by VWFS to a customer, is an inspection done by an appraiser to assess the damages. 
The appraiser 
identifies the damages by collecting: pictures of the damage, repair actions, and cost of these repairs. 
The process is a pain-point for the customers as well as VWFS because the customer can receive an unexpectedly high final bill while VWFS can receive vastly different appraisal for similar cars by different appraisers. 
In order to overcome this issue
we have leveraged the gains made by Neural networks for the task of Object detection. 

An issue with the data was the long tail distribution. 
The total types of damages in the dataset were 35, however, 84\% of the dataset can be captured by looking at the top 14 damage types, we have followed the same dataset treatment as \cite{b2} thereby removing damages such as "smells" and "missing" and only looking at the most frequent damages, following \cite{b2} 
two balanced datasets (D1,D2) are prepared for training. D1 comprised of the top 3 frequent damages and is  a subset of D2, which contained the top 9 frequent damages. 
We also added a background class to the original 9 classes to improve training.

For our model, we used Faster-RCNN \cite{b1}, with a ResNet50 pretrained feature extractor. 
The network is fine-tuned for 5 and 10 epochs for D1 and D2 respectively using an SGD optimizer. 
Our model reached a mean average precision (mAP) of 66.1\% for D1 and 49.3\% for D2. These inital results indicate that damage detection can be handled by our method and a more refined dataset will lead to consistent performance.

\section{Used-Cars Recommender system}

In VWFS, selling used cars is usually done after the end of the leasing contracts through online business-to-business auctions where dealers are presented with various options to choose from and bid on. 
To enhance the dealers' experience and improve the number of sales, a state-of-the-art attribute-aware multi-relational sequential model is developed by extending and adapting the recent SASRec \cite{kang2018self} model for auction settings \cite{rashed2020multirec}. The extended version SASRec-AUC uses the car configuration details as preprocessed numerical feature vectors, which replace the original one-hot encoded item ids as inputs in SASRec. Furthermore, inspired by our previous state-of-the-art auction-based model \cite{rashed2020multirec}, SASRec-AUC is trained using the two available historical relational interactions between user and items, namely the purchase and bidding interactions. Using both relations provides the model with richer data to learn dealers' preferences as both interactions are highly correlated.

To evaluate SASRec-AUC, we compared it against the state-of-the-art MultiRec model using the same evaluation protocol and VWFS dataset. In contrast to MultiRec, SASRec-AUC is a sequential attention-based model that can capture sequential patterns in dealers' histories, while MultiRec can not because it is a point-wise non-sequential recommender model. 

Results in Table \ref{UsedRec}, show that SASRec-AUC can significantly outperform MultiRec and other baseline models on both HitRatio and the Normalized Discounted Cumulative Gain (NDCG) metrics, especially when using all relations. Metrics concerning top 20 recommended items are reported as those align the most with business requirements.

\begin{table}[!ht]
\caption{Performance comparison on the VWFS dataset}
\label{UsedRec}
\begin{center}
  \begin{tabular}{|l|c|c|}
  \hline
    Model &HR@20&NDCG@20\\
    \hline
    Random  & 0.193&0.066\\
    top-Popular &0.388&0.260\\
    \hline
    MultiRec + Bidding + Sale Price & 0.657&0.366\\  
    \hline
    SASRec-AUC & 0.671&0.358\\
    SASRec-AUC + Bidding &\textbf{0.692} &\textbf{0.369}  \\
    \hline
  \end{tabular}
  \end{center}
\end{table}

\section{Next Best Offer}
The Next best offer work package can be formulated as a recommendation problem, where given a previous vehicle contract we aim to predict a personalized shortlist of renewal offers across customers. 

We consider 32,152 contracts dated from 2008 to 2020 from the Spanish market at the B2C transaction level. For learning, we use customer attributes (occupation, credit risk score etc.), context information (contract duration, geographic info. etc.) and vehicle attributes (model year, fuel type etc.)

We propose a customized deep embedding network for learning customer preferences, built with dedicated embedding components for all categorical variables. Numerical variables are embedded altogether with non-linear fully-conencted (FC) layers. Next, we concatenate these embeddings and apply a series of contraction operations through FC layers to learn hierarchical correlations between all categorical and numerical variables. Finally, a fully-connected layer outputs probabilistic scores (softmax activated) for each vehicle class as the next purchase, which can be used to generate a ranking. 

We experimentally validate the effectiveness of our proposed method by comparing it to a series of heuristic, Random, Repeat+Top-pop, Nearest Neighbours and a learned baseline approach CatBoost \cite{b4}. Repeat-pred+top-popular ensembles the previous vehicle and the most popular vehicle classes. Our previous work \cite{rashed2020multirec} and references therein elaborate more on the other baselines and metrics. Finally, the evaluation protocol divides the available contract data into 70\% training, 10\% validation and 20\% test splits randomly. HR@5 results are 0.07, 0.55, 0.41, 0.74 and 0.76 in order of baselines introduced and lastly the proposed method. Similarly, for NDCG@5, 0.04, 0.44, 0.28, 0.57 and 0.59. Results achieved through the two learned methods for HR@5 can enable various downstream B2C marketing possibilities. As before, we report metrics for top 5 recommendations that align with business requirements. 

\section{Future Work}
As a future work, it would be interesting to learn a single multi-task model to predict all the targets required for (I) `Asset return forecast' by capitalizing on a shared encoding for these targets \cite{b3}. For (II) `Lease Vehicle Return Assessment', we want to extend the detection model to be aware of car specifications and also predict cost of repair along with damage identification, thereby providing an end-to-end solution. For (III) `Used Car Recommender system', we are planning to extend the model for accommodating external contextual features and evaluating it using AB testing. Lastly, for (IV) `Next best offer', we aim to extend the modeling to fine-grained recommendations that allows predicting factors such as trim level and fuel type of next vehicles.

\appendices
\section{Asset Return Forecast}


We observe that the CatBoost model is able to outperform more powerful Neural Network architectures, justifying the use of such boosted trees as a primary baseline for similar tabular data tasks. Additionally, the chronological evaluation protocol showcased consistent results with how the models are to be utilized in the business process workflows.

\section{Leasing Vehicle Return Assessment }

As a rudimentary baseline and as a first interaction with the dataset, we experimented with an ImageNet pretrained InceptionV3 classifier. We were able to get a 35\% classification accuracy when using all 42 damage classes, to put it in perspective a random model would have and accuracy of 21\%. This experiment highlighted the shortcomings of the dataset such as poor image quality, poor context capture inside the images as well as background interference in the images, a sample of the dataset is presented in \ref{sample_dataset}.

\begin{figure}[ht]
\centering
\begin{minipage}[b]{0.30\linewidth}
\includegraphics[scale=0.06]{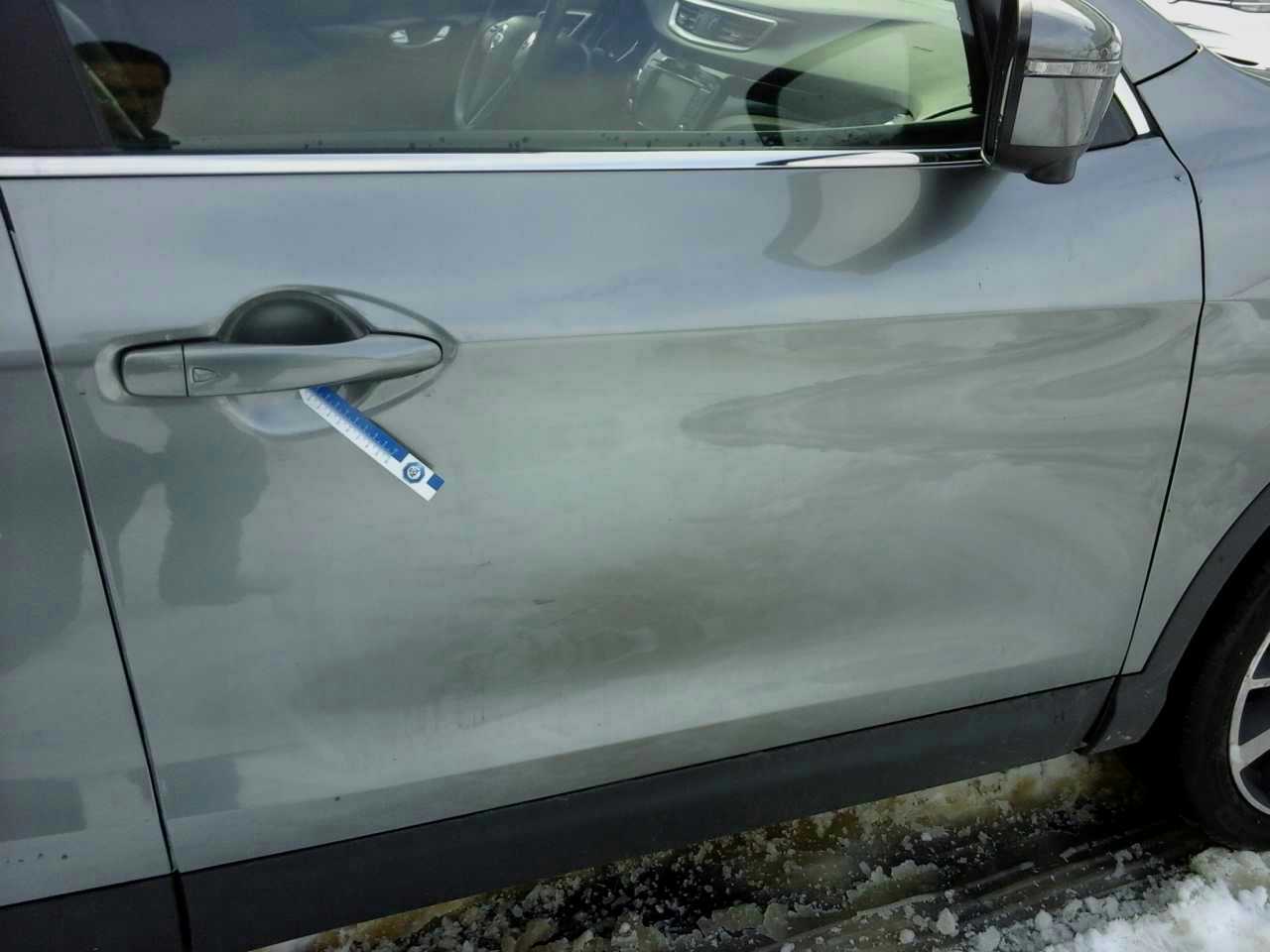}

\end{minipage}
\quad
\begin{minipage}[b]{0.30\linewidth}
\includegraphics[scale=0.06]{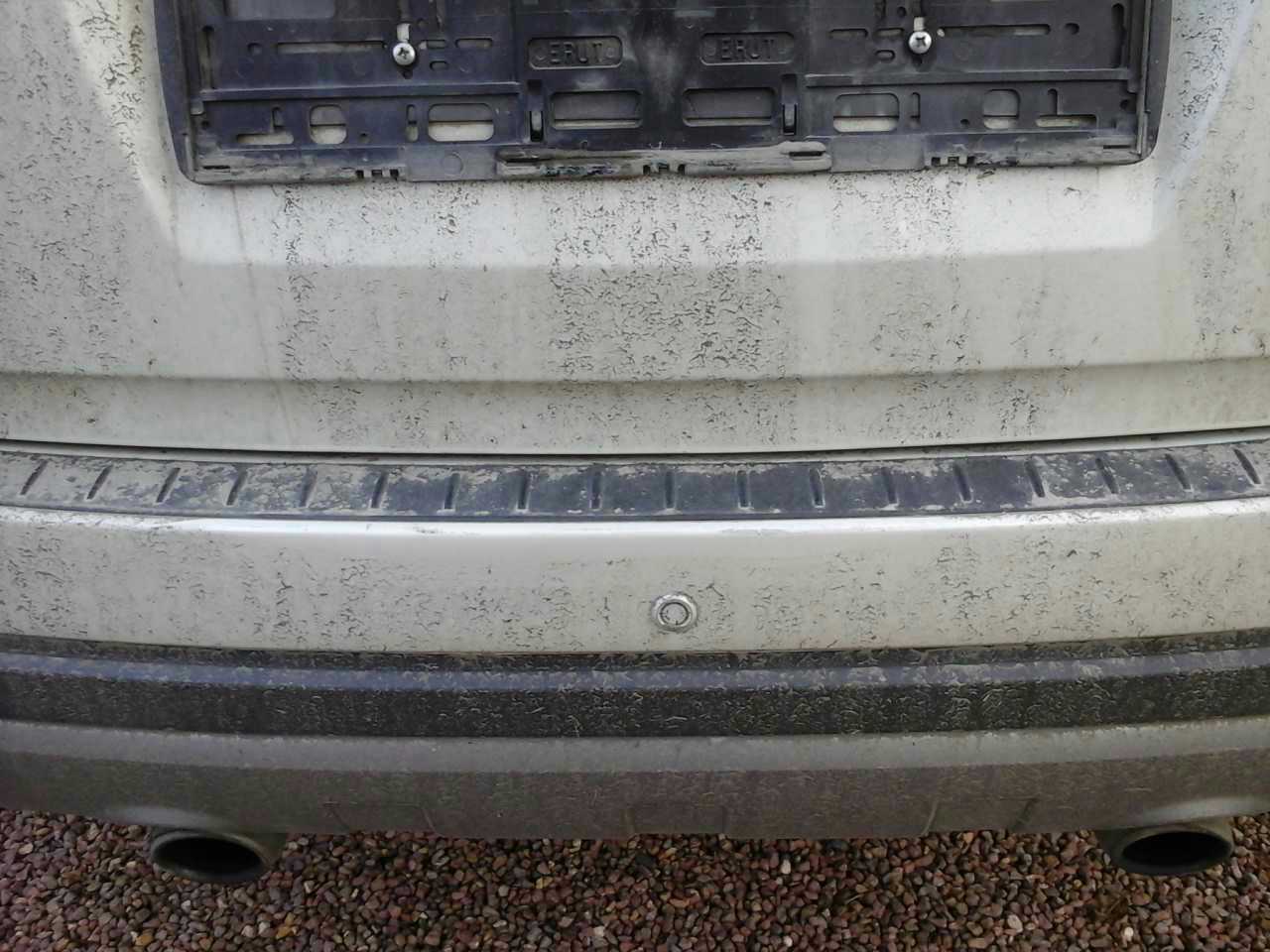}

\end{minipage}
\begin{minipage}[b]{0.30\linewidth}
\includegraphics[scale=0.06]{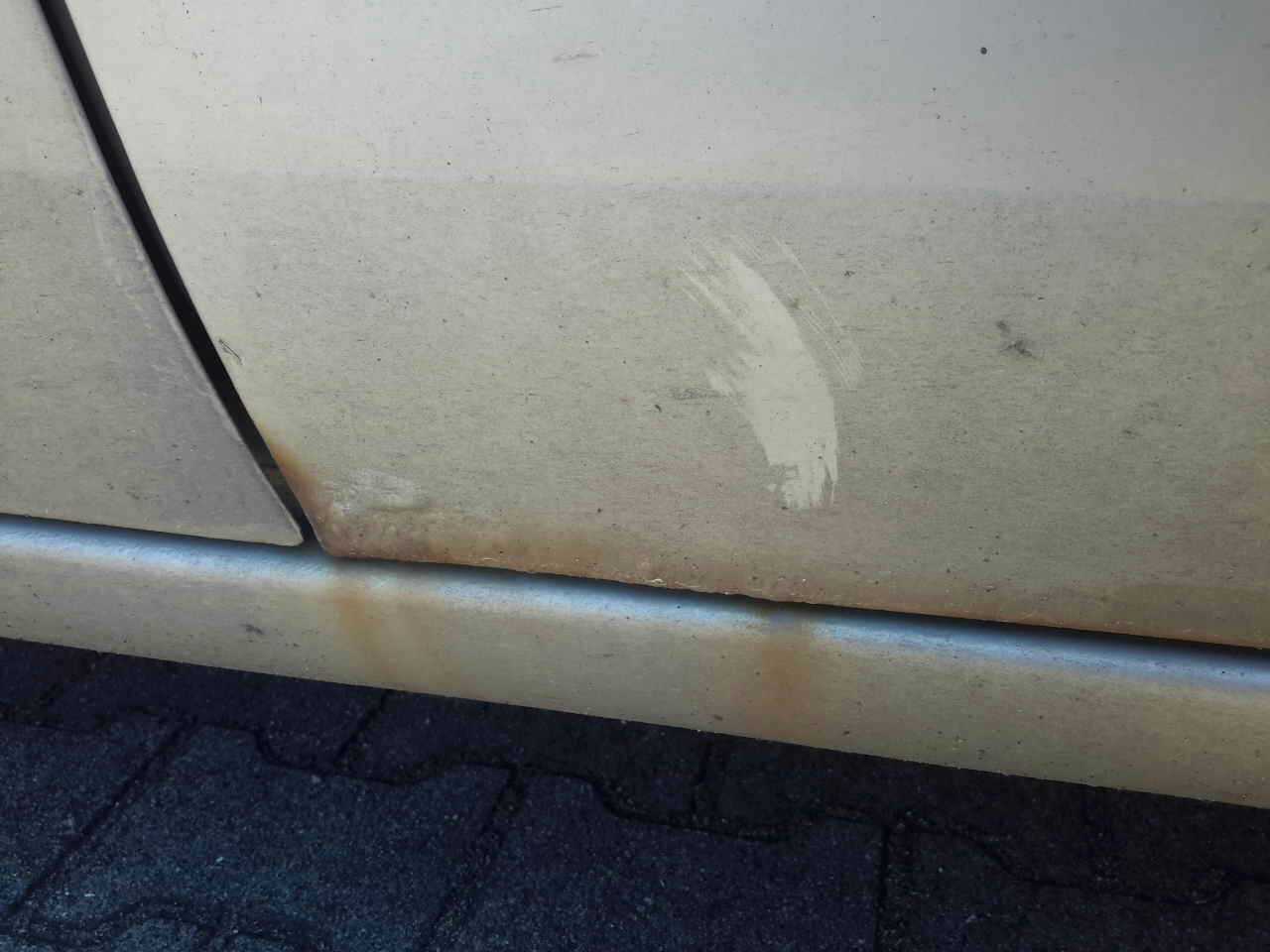}

\end{minipage}
\caption{Sample damage images }
\label{sample_dataset}
\end{figure}

In order to overcome the shortcomings of the dataset, we selected a subset of the damage types to include only the top 3 and top 10 damages (D1 and D2). This subset was further distilled to remove poorly taken pictures, we then annotated the dataset with bounding boxes. At the end we had 36000 total annotations from 50000 images. These annotated images were then used to get crops of the damages which were then used to train the InceptionV3 classifier, and we were able to get an accuracy of 93.4\% for the top 3 damage classes.

\begin{figure}[ht]
\centering
\begin{minipage}[b]{0.30\linewidth}
\includegraphics[height=0.8\linewidth]{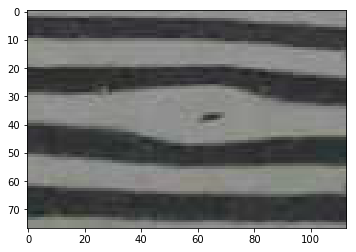}

\end{minipage}
\quad
\begin{minipage}[b]{0.30\linewidth}
\includegraphics[height=0.8\linewidth]{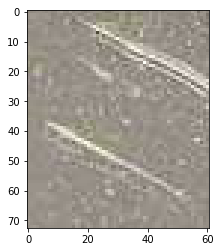}

\end{minipage}
\begin{minipage}[b]{0.30\linewidth}
\includegraphics[height=0.8\linewidth]{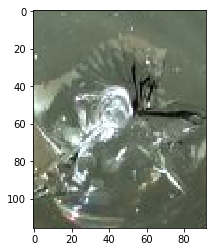}

\end{minipage}
\caption{Sample Annotated/cropped damages }
\label{sample_dataset}
\end{figure}

Having shown classification gains using bounding box damage crops, we have also presented the results on the object detection in section III.

The main lessons learned from this exercise were (i) The data collection needs to intentionally cater for the machine learning task, quality of the images play a part in the final model performance and (ii) Damage classes need to be defined in a way where the labels are distinct and unique, this will prevent `scratch' and `scratched' damage classes from appearing in datasets. 

\section{USED-CARS RECOMMENDER SYSTEM }
\begin{table}[]
\centering
{%
\caption{USED-CARS RECOMMENDER SYSTEM Datasets Statistics}
\label{b2b_dataset}
\begin{tabular}{|l|l|}
\hline
Dataset       & VWFS       \\ \hline
Users          & 3,220                  \\ \hline
Items            & 269,104                   \\ \hline
Purchases         & 269,104               \\ \hline
Biddings         & 375,349               \\ \hline
Purchases Density (\%)         & 0.031                \\ \hline
Biddings Density (\%)        & 0.043                \\ \hline
User Features         & 7,750                \\ \hline
Item Features         & 572                \\ \hline
Unique Items (\%)         & 99.999\%                \\ \hline

\end{tabular}%
}
\end{table}

The employed dataset in training the used-car recommender system is a Business-to-business (B2B) proprietary dataset that was collected by Volkswagen Financial Services used-cars center during the period between 2015 and 2019. The dataset includes historical purchase and bidding records from multiple brands and independent dealers. Table \ref{b2b_dataset} presents a summary of datasets statistics

The main lessons learned from implementing the used-car recommender systems can be listed as follows (i) Augmenting the data with available auxiliary user-item interactions, such as by using the bidding interactions besides the purchase interactions had a significant impact on the model's prediction accuracy and (ii) The threshold value and the number of negative items to be sampled in the evaluation protocol should be defined based on the expected page size that will be used to display the recommended items and based on the expected number of items available at one auction or session.

\section{Next Best Offer}
Our initial assumption was that a customized deep learning method would outperform the heuristic and other machine learned baselines significantly. However, in practice, we learned that model complexity and the engineering costs that ensue can be traded off for higher explainability and a slight decrease in performance. As a result the CatBoost model serves us as a viable option for future research into fine-grained recommendations and opens the door for sister models in the decision tree model family.

\end{document}